%% file: main.tex
\documentclass[10pt,twocolumn,letterpaper]{article}

\usepackage[pagenumbers]{cvpr} 

\input{preamble}

\definecolor{cvprblue}{rgb}{0.21,0.49,0.74}
\usepackage{multirow}
\usepackage[pagebackref,breaklinks,colorlinks,citecolor=cvprblue]{hyperref}
\usepackage[capitalize]{cleveref}
\usepackage{utfsym}
\usepackage{bm}
\crefname{section}{Sec.}{Secs.}
\Crefname{section}{Section}{Sections}
\Crefname{table}{Table}{Tables}
\crefname{table}{Tab.}{Tabs.}
\def\paperID{811} 
\def\confName{CVPR}
\def\confYear{2024}
\newcommand{\shortname}{\textsc{GVdiff}}
\title{\shortname{}: Grounded Text-to-Video Generation with Diffusion Models}

\author{Huanzhang Dou \quad  Ruixiang Li \quad Wei Su \quad Xi Li\\
College of Computer Science \& Technology, Zhejiang University
}

\begin{document}
\twocolumn[{%
	\maketitle
	\renewcommand\twocolumn[1][]{#1}%
	\begin{center}
		\centering
            \includegraphics[width=\textwidth]{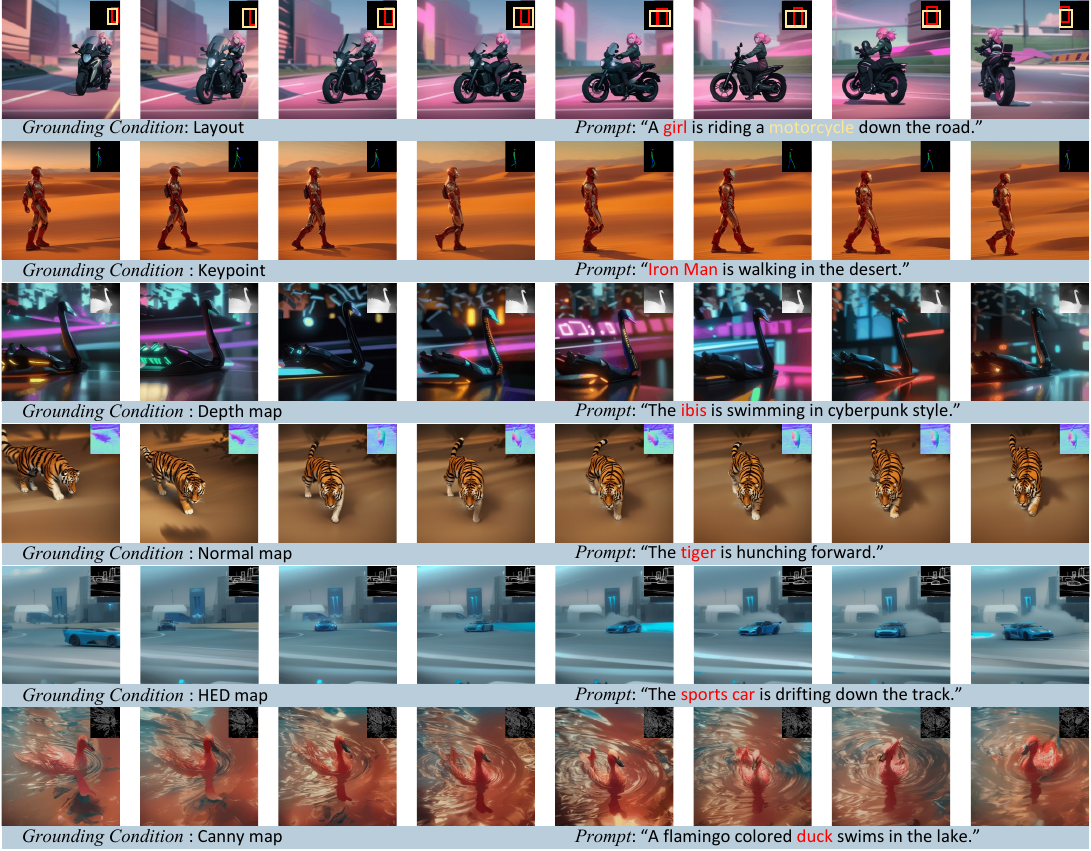}
		\captionof{figure}{\textbf{G}rounded text-to-\textbf{V}ideo (\shortname{}) generation aims to integrate text-to-video generation with grounded generation capacity under both discrete and continuous grounding conditions, including layout, keypoint, depth map, normal map, HED map, and canny map, \etc.}
		\label{fig:first}
	\end{center}
}]

\input{sec/0_abstract}    
\input{sec/1_intro}

\input{sec/2_related}

\input{sec/3_method}

\input{sec/4_exp}

\input{sec/5_conclusion}

{\small
\bibliographystyle{ieeenat_fullname}
\bibliography{main}
}
\newpage
\newpage
\appendix

\section{Additional Comparison with state-of-the-art Methods.}

To comprehensively demonstrate the effectiveness of~\shortname{}, we extend our analysis with additional comparisons to contemporary state-of-the-art methods. These comparisons, depicted in~\cref{fig:1} through~\cref{fig:5}, encompass a range of advanced techniques including Control-A-Video (CAV)~\cite{chen2023control}, ControlVideo (CV)~\cite{zhang2023controlvideo}, and Gen-1~\cite{esser2023structure}. 
\input{sec/comp1}

\section{Results with Personalized Models.}
Inspired by~\cite{guo2023animatediff,liew2023magicedit}, we use the multi-stage training paradigm to train the spatial-temporal grounding attention, dynamic gate network, and temporal attention. Therefore, each module focuses on its function alone rather than interfering with each other. We validate the effectiveness of \shortname{} on the popular personalized models from CivitAI~\cite{civitai} conditioned on web videos as shown in~\cref{fig:per}.
\input{sec/per}
\end{document}

%% file: preamble.tex
\usepackage[dvipsnames]{xcolor}


%% file: sec/0_abstract.tex
\begin{abstract}
In text-to-video (T2V) generation, significant attention has been directed toward its development, yet unifying discrete and continuous grounding conditions in T2V generation remains under-explored. This paper proposes a \textbf{G}rounded text-to-\textbf{V}ideo generation framework, termed \shortname{}. First, we inject the grounding condition into the self-attention through an uncertainty-based representation to explicitly guide the focus of the network. Second, we introduce a spatial-temporal grounding layer that connects the grounding condition with target objects and enables the model with the grounded generation capacity in the spatial-temporal domain. Third, our dynamic gate network adaptively skips the redundant grounding process to selectively extract grounding information and semantics while improving efficiency. We extensively evaluate the grounded generation capacity of~\shortname{} and demonstrate its versatility in applications, including long-range video generation, sequential prompts, and object-specific editing.

\end{abstract}

%% file: sec/1_intro.tex
\section{Introduction}
\label{sec:intro}

Text-to-image (T2I) generation has witnessed significant success, propelled by GAN~\cite{zhang2017stackgan, xu2018attngan, reed2016generative, zhu2019dm, goodfellow2014generative} and diffusion models (DMs)~\cite{feng2023ernie,zhou2022towards,gu2022vector,li2022upainting}. Pioneering efforts such as DALL-E2~\cite{esser2023structure}, Imagen~\cite{saharia2022photorealistic}, Cogview~\cite{ding2021cogview}, and Latent Diffusion~\cite{rombach2022high} have set new benchmarks. Besides, controllable technology~\cite{li2023gligen, Zhang_2023_ICCV,mou2023t2i} has expanded the horizons of generation, allowing for precise manipulation in image composition under specific conditions.

Building upon the above progress, text-to-video (T2V) generation~\cite{wu2022nuwa,ho2022imagen,blattmann2023align} has gained increasing attention. T2V generation poses challenges such as data scarcity, complex temporal dynamics, and substantial computational demands. Approaches like CogVideo~\cite{hong2022cogvideo} and Make-A-Video~\cite{singer2023makeavideo} utilize T2I models by freezing or fine-tuning their weights. Then, Control-A-Video~\cite{chen2023control} and ControlVideo~\cite{zhang2023controlvideo} leverage ControlNet~\cite{Zhang_2023_ICCV} to achieve controllable T2V generation. However, unifying discrete grounding (\eg., bounding box) and continuous grounding (\eg., canny map) in T2V generation remains under-explored.

In this paper, we introduce a framework termed \textbf{G}rounded text-to-\textbf{V}ideo generation (\shortname{}), designed to facilitate grounded T2V generation under discrete and continuous grounding conditions. \shortname{} harnesses pre-trained T2I models to capitalize on their established proficiency in photorealistic content generation and language understanding, thereby circumventing the need for extensive training from scratch. Specifically, we inject grounding conditions into the self-attention mechanism by transforming them into uncertainty-based representations, which could explicitly direct the focus of the network. We then empower the grounded generation capacity via an introduced spatial-temporal grounding layer (STGL), which is engineered to connect the grounding condition with the targeted object, facilitate the interaction between grounded features and visual tokens, and ensure temporal consistency. Further, we propose a dynamic gate network (DGN) that selectively bypasses redundant grounding operations in STGL to improve efficiency since shallow and deep layers are responsible for dealing with low-level details (\eg., grounding information) and high-level semantics respectively.

Our experimental evaluation of \shortname{} is conducted under both discrete and continuous grounding conditions as depicted in~\cref{fig:first}—attest to its capability to generate videos with high temporal consistency and precise grounded control. With its grounded generation capacity, \shortname{} demonstrates three practical applications, including long-range video generation, sequential prompts, and object-specific editing.

The main contributions are summarized as follows:
\begin{itemize}
    \item We propose a \textbf{G}rounded text-to-\textbf{V}ideo Generation (\shortname{}) framework. This approach incorporates a proposed spatial-temporal grounding layer, enabling both discrete and continuous grounding control over the video generation process.
    \item We propose a dynamic gate network (DGN), which aims to adaptively bypass redundant grounding operations to selectively process low-level grounding information and high-level semantics while improving efficiency.
    \item We evaluate the effectiveness of grounded T2V generation capacity of \shortname{}. Further, \shortname{} has three practical applications, including long-range video generation, sequential prompts, and object-specific editing.
\end{itemize}

%% file: sec/2_related.tex
\section{Related Work}

\subsection{Diffusion Model} The generative AI~\cite{cao2023comprehensive,zhang2023complete,Su2023ReferringEC,Su_2024_CVPR,su2023language} has undergone rapid development from the generative adversarial network (GAN)~\cite{goodfellow2014generative} to the latest diffusion models (DMs)~\cite{ho2020denoising}. Denoising diffusion probabilistic model (DDPM)~\cite{ho2020denoising, sohl2015deep} has achieved promising
success in generation and editing tasks, such as text-to-image (T2I), text-to-speech~\cite{chen2020wavegrad, chen2021wavegrad, kong2021diffwave}, text-to-3D~\cite{watson2022novel}, and text-to-video (T2V)~\cite{hong2022cogvideo,wu2023tune,he2022latent,ho2022imagen}. Especially in T2I task, the latent diffusion model (LDM)~\cite{rombach2022high} and the successor Stable Diffusion (SD) are at the forefront of generative research.  LDMs consist of two main components, \ie, autoencoder and DDPM, which aims to remove the noise added to the sampled data. The DDPM is implemented via UNet~\cite{ronneberger2015u}, which takes a noised latent $\bm{z}$, time step $t$, and caption $\bm{c}$ as input. Besides, a BERT-like~\cite{devlin2018bert} network encodes each text description into a sequence of text embeddings. Specifically, the caption feature is derived from a frozen text encoder of the CLIP ViT-L/14~\cite{radford2021learning} in Stable Diffusion. The caption $\bm{c}$ interacts with UNet through the cross-attention layer.

The diffusion forward process commences with an image $\bm{x}_0$, which is transformed into $\bm{z}_0=\mathcal{E}(\bm{x}_0)$ by a frozen encoder $\mathcal{E}(\cdot)$ Subsequently, a predefined Markov process introduces noise perturbation until reaching $\bm{z}_T$:
\begin{equation}
    q(\bm{z}_t|\bm{z}_{t-1})=\mathcal{N}(\bm{z}_t;\sqrt{1-\beta_{t-1}}\bm{z}_{t-1}, \beta_{t-1} I).
\end{equation}

The objective of DMs is to learn the denoising process (reversed diffusion). Given a sampled random noise $\bm{z}_t$, the model learns to predict the added noise at the next timestep $\bm{z}_{t-1}$ until $\bm{z}_0$:
\begin{equation}
    p_{\theta}(\bm{z}_{t-1}|\bm{z}_t)=\mathcal{N}(\bm{z}_{t-1};\mu_{\theta}(\bm{z}_t,t),\Sigma_\theta(\bm{z}_t,t)).
\end{equation}

LDM enhances computational and memory efficiency over pixel-space diffusion models through a two-stage training paradigm. First, LDM trains an autoencoder to map the input image into a spatially low-dimensional latent space of reduced complexity. The encoder $\mathcal{E}(\cdot)$  and decoder $\mathcal{D}(\cdot)$ reconstructs input $\bm{x}$ with $\hat{\bm{x}}=\mathcal{D}(\mathcal{E}(\bm{x}))\approx \bm{x}$.  Second, the diffusion model is trained on the latent $\bm{z}$. During the training process, LDM $f_{\theta}$ is optimized to gradually remove the noise $\varepsilon$, which is added to the latent $\bm{z}_t$:
\begin{equation}
\min\mathcal{L}_{LDM}=\mathbb{E}_{\bm{z},\varepsilon\sim\mathcal{N}(0,I),t,\bm{c}}[\Vert \varepsilon-f_{\theta}(\bm{z}_{t},t,\bm{c})\Vert^2_{2}],
\end{equation}
where $t$ is uniformed sampled from the time steps, mapped to time embedding $\phi(t)$, and then fed into the UNet.

\subsection{Text-to-Video Models}
\textbf{Text-to-Video Generation.} Existing approaches aim to leverage transformers and diffusion models for video generation~\cite{yu2023video, luo2023videofusion,singer2023makeavideo,zhou2022magicvideo,chen2023executing,ni2023conditional,villegas2023phenaki}. NUWA~\cite{wu2022nuwa} introduces a 3D transformed encoder-decoder framework, supporting both T2I and T2V generation. Imagen Video~\cite{ho2022imagen} constructs spatial-temporal super-resolution models to generate temporally consistent video with high resolution. AnimateDiff~\cite{guo2023animatediff} injects a motion module to animate off-the-shelf T2I models without the need for model-specific tuning.

\noindent\textbf{Text-to-Video Editing.} Compared to image editing, video editing~\cite{vid2vid-zero,wang2023gen,couairon2023videdit,qi2023fatezero,geyer2023tokenflow,zhou2022magicvideo} is more challenging for geometric and temporal consistency. Tune-A-Video~\cite{wu2023tune} proposes one-shot video editing by extending and tuning T2I models on a single reference video. Dreammix~\cite{molad2023dreamix} develops a text-to-video backbone from motion editing while preserving temporal consistency with high fidelity. Fate-zero~\cite{qi2023fatezero} and Text2Video-zero~\cite{khachatryan2023text2video} explore generate videos only using pre-trained T2I models. Layered neural atlases~\cite{bar2022text2live,kasten2021layered,chai2023stablevideo} address editing by decomposing the video into a set of unified 2D atlases layer for each target, allowing contents to be applied to the global summarized 2D atlases and mapping back to the video with temporal consistency.

\subsection{Grounded Diffusion Models} 
Grounding conditions are primarily divided into discrete (\eg., layout) and continuous (\eg., depth map) conditions.

For discrete conditions, prior works \cite{yang2023reco,yang2022modeling,sylvain2021object,sun2021learning,li2021image,jahn2021high} aim to generate images from bounding boxes labeled with object descriptions,  performing the inverse of object detection \cite{carion2020end}.  Recently, zero-shot layout2image generation~\cite{chen2023training,couairon2023zero} is proposed to save the training costs. Text2layout~\cite{qu2023layoutllm,zhang2023controllable} is also proposed to add more spatial prior. GLIGEN~\cite{li2023gligen} extends the grounded entities to be open-world. Then, MOVGAN~\cite{wu2023multi} pioneers the extension 
 from layout2image generation to layout2video generation but only with the layout of the first frame, limiting the controllability over the entire video.

For continuous conditions, ControlNet~\cite{Zhang_2023_ICCV} and T2I-adaptor~\cite{mou2023t2i} add spatial grounding conditions to T2I DMs by introducing additional modules to represent the conditions, such as canny map, depth map, and human pose. Then,  Control-A-Video~\cite{chen2023control} and ControlVideo~\cite{zhang2023controlvideo} extend ControlNet to enable controllable video generation though they fall short in addressing discrete grounding conditions.

In this paper, we introduce a general framework to incorporate grounding conditions for T2V generation. Further, based on the grounded generation capacity, we develop three practical applications, including long-range video generation, sequential prompt, and object-specific editing.

%% file: sec/3_method.tex
\input{img_tex/overview}

\section{Method}

The overview of \shortname{} is illustrated in~\cref{fig:overview}. In this section, we elucidate the modified self-attention with uncertainty-based grounding injection, the spatial-temporal grounding layer, and the dynamic gate network. Then, we introduce three practical applications with the capacity of grounded T2V generation. We take layout-conditioned T2V generation for example if there are no special instructions, owing to its flexibility in adapting to continuous conditions

\subsection{Uncertainty-based Grounding Injection} Grounded video generation strives for enhanced controllability with additional grounding conditions. Previous methods~\cite{chen2023control,li2023gligen} overlook the injection of grounding conditions during the feature extraction phase, which makes the visual token integration process unaware of the grounding conditions. To address this, we propose an uncertainty-based grounding injection for the self-attention of UNet to explicitly constrain the focused region of the network.

Specifically, we enhance the attention region with grounding conditions via a non-parametric Gaussian transform. Given a grounding condition $\bm{l}=[\bm{l}_0,\dots,\bm{l}_{M-1}]$ with $M$ target objects, the coordinate value of corresponding two-dimension Gaussian probability distribution $\bm{M}_k$ of each object could be formulated as:

\begin{equation}
    \bm{M}_k(i,j) = \mathrm{Gauss}(\bm{l}_k)=\frac{1}{2\pi\sigma_{x_k}\sigma_{y_k}}e^{-(\frac{(i-\mu_{x_k})^2}{2\sigma_{x_k}^2}+\frac{(j-\mu_{j_k})^2}{2\sigma^2_{y_k}})}.
\end{equation}

The averaged uncertainty-based grounding map $\bm{M}=\frac{1}{M}\sum_{k=0}^{M-1}\bm{M}_k$ is injected into the self-attention layer of the UNet to guide the visual token integration as:
\begin{equation}
     \bm{z}=\mathrm{SelfAttn}(\bm{Q},\bm{K},\bm{V},\bm{M})=\mathrm{Softmax}(\frac{\bm{Q}\bm{K}^T}{\sqrt{d}}+\bm{M})\cdot \bm{V}.
\end{equation}

This strategy is applied to sparse grounding conditions, including layout, keypoint, and depth. For dense grounding conditions, we leverage Gaussian blur to emphasize the prior.

\subsection{Spatial-Temporal Grounding Layer}

Training a spatial-temporal UNet from scratch has the following drawbacks. First, 3D ResNet and Transformer introduce heavy parameter/computation costs. Second, since pre-trained Stable Diffusion contains abundant content synthesizing and language understanding capability, training another model is inefficient. Besides, finetuing diffusion model may cause catastrophic forgetting issues~\cite{ruiz2023dreambooth}.

To empower grounded generation capacity to the T2V model while improving training efficiency, we modify the transformer layer of pre-trained UNet with the proposed spatial-temporal grounding layer (STGL) to compose Grounded-UNet, instead of training from scratch. Inspired by~\cite{guo2023animatediff}, this training paradigm has the potential to be applied to any personalized models, \ie., DreamBooth~\cite{ruiz2023dreambooth} or LoRA~\cite{hu2021lora} models.

After using grounding condition to guide the focus region of self-attention, the relation between target objects and corresponding grounding conditions is established as~\cite{li2023gligen}. Given $M$ target objects $\bm{e}=[\bm{e}
_0^0,\dots,\bm{e}_{M-1}^{N-1}]$ and its corresponding grounding conditions $\bm{l}=[\bm{l}_0^0,\dots, \bm{l}_{M-1}^{N-1}]$ with $N$ frames, the grounded features $\bm{g}=[\bm{g}^0_0,\dots, \bm{g}^{N-1}_{M-1}]$ could be obtained following~\cite{li2023gligen}:

\begin{equation}
    \bm{g}_i^j=\mathrm{MLP}([f_{text}(\bm{e}_i^j), \mathrm{Fourier}(\bm{l}_i^j)]),
\end{equation}
where $\mathrm{Fourier}$ refers to the Fourier embedding~\cite{mildenhall2020nerf} and $\mathrm{MLP}(\cdot)$ is a multi-layer perceptron to align the feature dimension. The features of the target objects are extracted by the text encoder $f_{text}$ of CLIP~\cite{radford2021learning}.

Simply introducing grounded T2I methods into T2V generation is suboptimal. For grounded T2I generation~\cite{li2023gligen}, only one frame should be conditioned on the grounding condition. However, the content of every frame of the generated video using grounded T2I generation may adhere to the grounding condition but completely temporal discrete, particularly under sparse grounding conditions like layout and depth maps. Therefore, we plug temporal attention atop the grounded features to enhance their temporal consistency and then perform the spatial-temporal grounding attention (STGA) in~\cite{li2023gligen} as:

\begin{equation}
    \bm{z}=\bm{z}+\beta \cdot\mathrm{tanh}(\gamma)\cdot \mathrm{TS}(\mathrm{SelfAttn}([\bm{z},\mathrm{TempAttn}(\bm{g})])),
\end{equation}
where $\gamma$ is a learnable parameter and $\beta$ affects the controllability. $\mathrm{TS}$ signifies the token selection operation to consider visual tokens and exclude grounded tokens. Then, caption feature $\bm{c}$  interacts with the visual tokens via cross attention:

\begin{equation}
    \bm{z} = \bm{z} + \mathrm{CrossAttn}(\bm{z}, \bm{c}).
\end{equation}

Following the mainstream practice~\cite{guo2023animatediff,blattmann2023align} to improve temporal consistency, we introduce an additional temporal attention layer. This layer learns to align each frame within a temporal consistent paradigm. We adopt a zero initialization strategy to prevent disturbances in the initial feature distribution of the output. The architecture of STGL is illustrated in~\cref{fig:overview}(c).

\subsection{Dynamic Gate Network}

After enhancing the T2V model with the capacity of grounded generation, an important question emerges: Is it essential for grounding conditions to engage with visual tokens at every layer? This consideration stems from the understanding that low-level and high-level information is predominantly processed by the shallow and deep layers\cite{zhang2022contrastive, Pang_2019_ICCV, He_2021_ICCV, yuan2024semanticmimmarringmaskedimage}, respectively. Therefore, excessive grounding interaction can not only impede grounding/semantic extraction but also diminish overall efficiency.

To tackle this issue, we propose the Dynamic Gate Network (DGN), which could adaptively bypass the spatial-temporal grounding attention upon specific grounding conditions. We establish layer-wise grounding-aware embedding $\bm{v}_i$ to represent the $i^{th}$ layer's sensitivity to specific grounding conditions. Then, the relevance score is deduced through token-wise attention by the inner product of $\bm{v}_i$ and pooled grounded features $\bm{p}=\frac{1}{N}\sum_{j=0}^{N-1}\bm{g}^j$. Each attention weights $\bm{\alpha}_i$ of layer $i$ are derived as $\bm{\alpha}_i=\mathrm{Softmax}([\bm{v}_i\cdot \bm{p}_0,\dots, \bm{v}_i\cdot\bm{p}_{M-1}])$. Then, the weighted sum of $\bm{p}$ is fed to the two layer low-rank MLP to produce the relevance $\bm{r}_i$ between $\bm{p}$ and $\bm{v}_i$. This relevance signifies the likelihood of the necessity of the grounded feature at the current layer.

Then, we employ the Gumbel-Softmax~\cite{jang2017categorical} to calculate the probability of gate activation while maintaining differentiability. Considering the binary decision, we replace the Softmax operation with Logistic. Besides, we use dual gate mechanisms (\ie., hard and soft) to stabilize the training, inspired by~\cite{kaiser2018discrete}. For the $i^{th}$ layer, the dual gate is:

\begin{equation}
      \bm{d}_i^s=\mathrm{Sigmoid}(\hat{\bm{r}_i}),  \,\,\,\bm{d}^h_i=\mathcal{I}(\hat{\bm{r}}_i\geq 0),
\end{equation}
\begin{equation}
     \bm{d}_i=\mathcal{I}(n_i\geq 0.5)\cdot\bm{d}^s_i + \mathcal{I}(n_i<0.5)\cdot \bm{d}^h_i,
\end{equation}
where $\hat{\bm{r}}_i=\bm{r}_i+\epsilon$, with $\epsilon$ sampled from $\mathrm{Logistic}(0,1)$. $\mathcal{I}(\cdot)$ is the indicator function. $\bm{d}_i^s\in [0,1]$ and $\bm{d}_i^h\in\{0,1\}$ correspond to the soft gate and hard gate of layer $i$, respectively. During training, $n_i$ is randomly sampled from $\mathrm{Uniform}(0,1)$ to determine the activation of either the soft or hard gate at layer $i$. The architecture is depicted as~\cref{fig:gate}.

\input{img_tex/gate}

\subsection{Applications}
Based on the grounded generation capacity of \shortname{}, we have extended our method to three practical applications.

\noindent\textbf{Long-range Generation.} Most of T2V methods~\cite{wu2023tune,singer2023makeavideo} typically generate 8-24 frames, which may be limited by hardware and software optimization. Our method predicts future frames conditioned on specific window sizes of preceding context frames. This approach theoretically facilitates video generation of infinite length through an auto-regressive process, amalgamating discrete video segments via context frames.

\noindent\textbf{Non-uniform Sequential Prompt.} To meet the increasing demand for longer videos, a single textual prompt is often insufficient for generating satisfactory results. Therefore, we propose using non-uniform sequential prompts, thereby enabling fine-grained control over the content generation process. Sequential prompts are injected into specific frames, supplemented by the interpolation of adjacent prompts for intermediary frames to ensure the semantic consistency of the generated video.

\noindent\textbf{Object-specific Editing.} Capitalizing on the grounded generation capacity of \shortname{} and advanced grounding methods , such as SAM~\cite{Kirillov_2023_ICCV} and GroundingDINO~\cite{liu2023grounding}, we facilitates object-specific editing while maintaining background consistency. We use SAM with a textual prompt to segment the target object, allowing the generated content to be restricted to the mask region. The edited video is composed of the inpainted background and the generated object.

%% file: img_tex/overview.tex
 \begin{figure*}[t]
    \begin{center}
       \includegraphics[width=\textwidth]{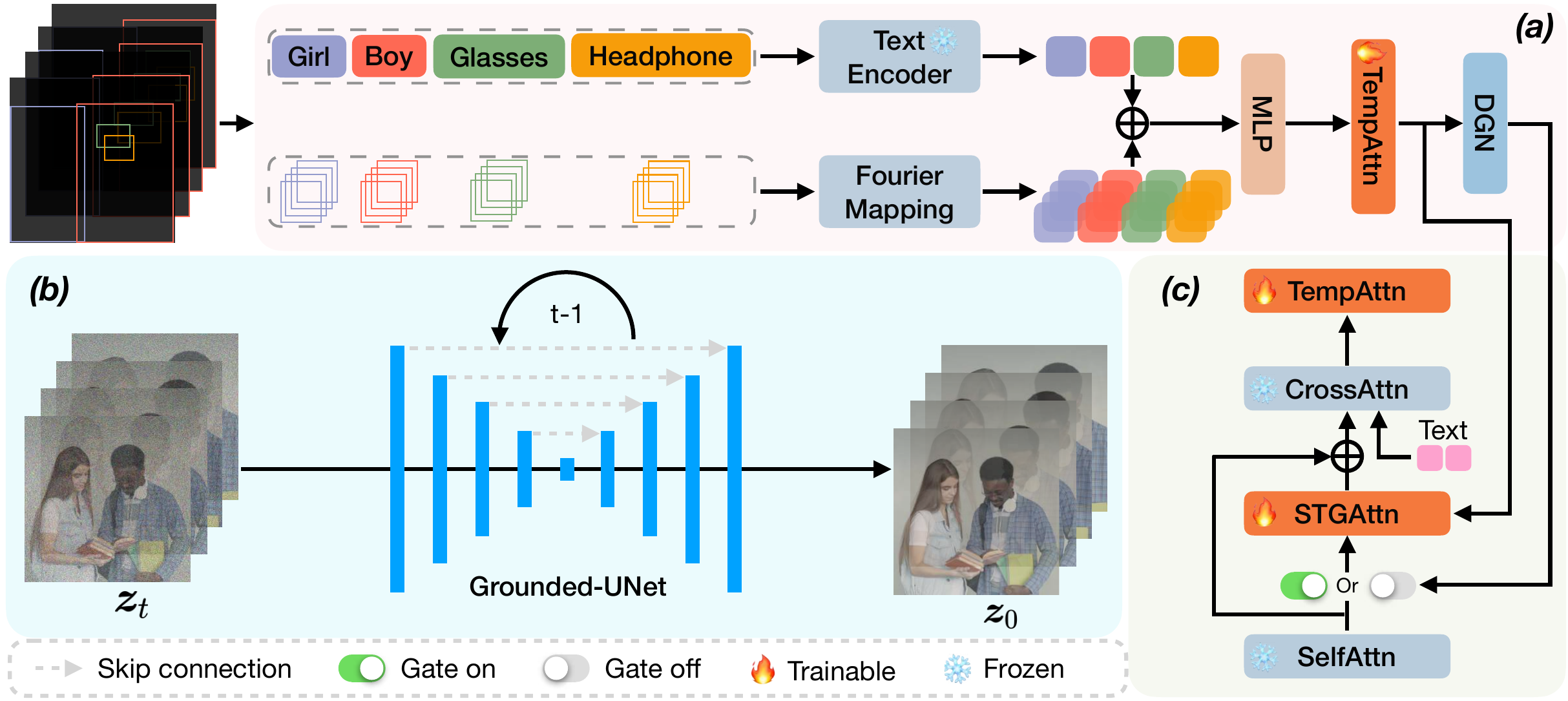}
    \end{center}
    \caption{Overview of \shortname{}. \textbf{\textit{(a)}} Connecting grounding conditions with target objects into grounded features, which are then smoothed by temporal attention. \textit{\textbf{(b)}} Generation with Grounded-UNet, where the transformer layer of UNet is replaced with the following spatial-temporal grounding layer. \textbf{\textit{(c)}} Spatial-temporal grounding layer. First, the uncertain-based grounding is injected into the self-attention. Then, spatial-temporal grounding attention (STGA) facilitates the interaction between the grounded features and visual tokens. An additional temporal attention layer ensures temporal consistency. Dynamic Gate Network (DGN) adaptively skips the redundant STGA.}
    \label{fig:overview}
 \end{figure*}

%% file: img_tex/gate.tex
 \begin{figure}[ht]
    \begin{center}
       \includegraphics[width=0.48\textwidth]{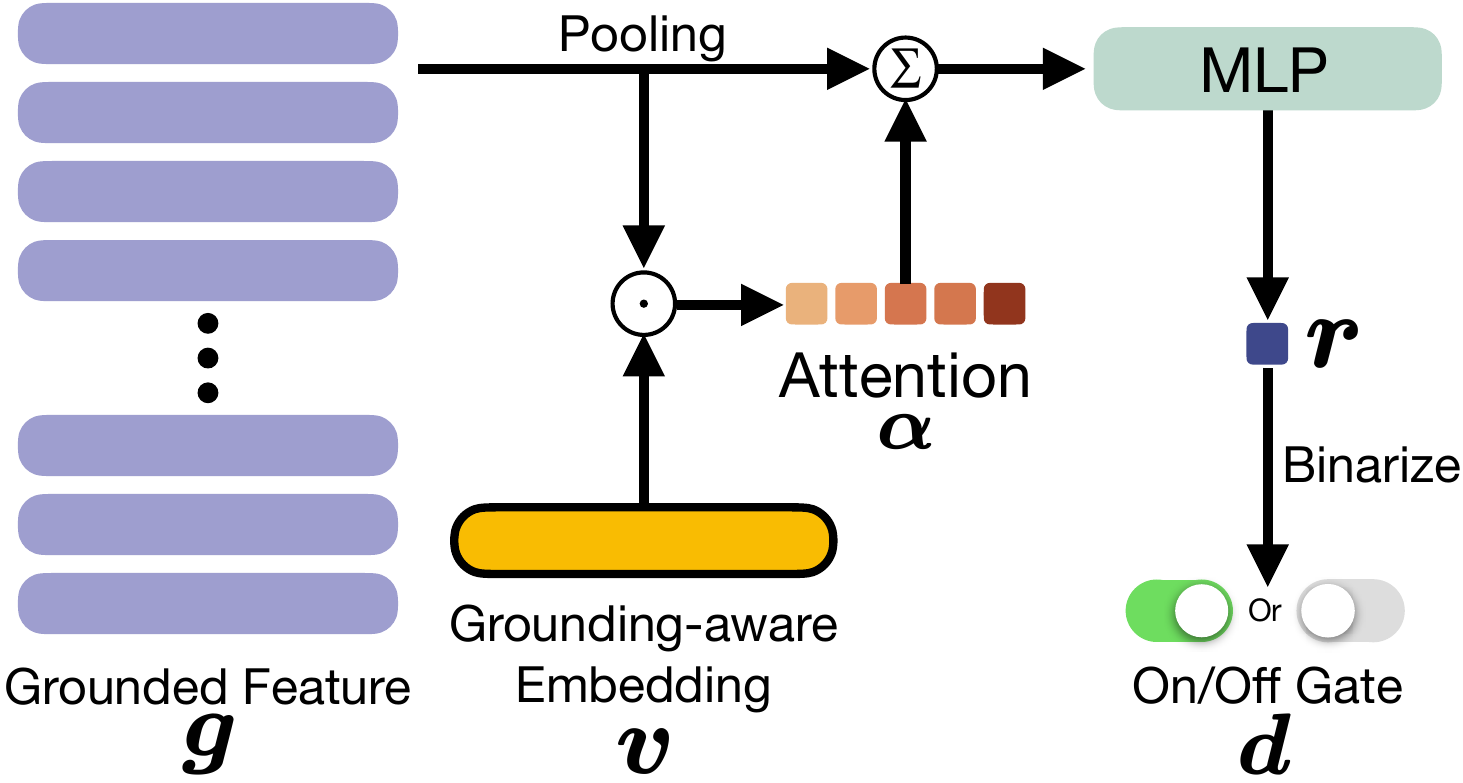}
    \end{center}
    \caption{Illustration of Dynamic Gate Network (DGN), which adaptively skips redundant spatial-temporal grounding attention. }
    \label{fig:gate}
 \end{figure}

%% file: sec/4_exp.tex
\section{Experiments}
\input{img_tex/main}

\subsection{Implementation Details}
\noindent\textbf{Training.} We utilize GLIGEN~\cite{li2023gligen}, a powerful grounded T2I model as our baseline. We train \shortname{} on WebVid-2M~\cite{bain2021frozen}. The video clips are sampled with 16 frames at the stride of 4. Missing objects and grounding are represented with the corresponding learnable null embeddings. \shortname{} is initialized the pre-trained models from Stable Diffusion v1.5, AnimateDiff, and GLIGEN. The frame-wise caption and layout are obtained by BLIPv2~\cite{li2023blip} and GLIPv2~\cite{zhang2022glipv2}, respectively. The other continuous grounding conditions are extracted by its corresponding detector. We use the multi-stage training paradigm to train spatial-temporal grounding attention, temporal attention, and the dynamic gate network. For the dynamic gate network, the soft gate is only to stabilize training and is discarded at the inference stage.

\noindent\textbf{Evaluation.} We use 25-step DDIM sampler and utilize classifier-free guidance~\cite{ho2022classifier} with the guidance scale of 7.5. We use a linear beta scheduler. Following the popular protocol~\cite{chen2023control,vid2vid-zero}, we conduct the comparison on 200 videos from DAVIS~\cite{ponttuset20182017} and web, which are cropped and resized into the square format for better visualization. We also validate the grounded generation capacity on personalized models in the supplementary material. For quantitative comparison, we use two metrics to evaluate the generation quality. First, Temporal Consistency is derived from the average cosine similarity between all pairs of the CLIP embedding of neighborhood frames. Second, Prompt Consistency is obtained by computing the average CLIP Score between generated videos and the driven text $p$.

\subsection{Main Results}
\noindent\textbf{Grounded Generation Capacity.} We qualitatively evaluate the grounded generation capacity of \shortname{} under discrete and continuous grounding conditions as illustrated  in~\cref{fig:first}, which indicates that \shortname{} could effectively generate high-fidelity and temporally consistent videos under both discrete and continuous grounding conditions. First, the well-trained T2I model ensures the generation of photorealistic content. Second, the proposed spatial-temporal grounding layer could precisely control the position of generated content and improve temporal consistency. Third, with the increasing details of grounding conditions ranging from the sparse layout to the dense canny map, the controllability becomes more fine-grained. 

Further, we quantitatively evaluate the grounded generation capacity shown in~\cref{tab:grounded}. For discrete layout, we compute AP following GLIGEN~\cite{li2023gligen}. For other continuous grounding conditions,  we compute averaged CLIP similarity between the condition of the source and generated videos to validate whether the grounding information in the generated video meets expectations. All conditions are extracted by the corresponding extractor. Results indicate that \shortname{} could effectively outperform previous grounded T2V methods.

\input{tab/ground}

\noindent\textbf{Qualitative comparison.} In a qualitative comparison with current state-of-the-art methods, presented in~\cref{fig:main}, \shortname{} demonstrates superior capability in generating content-rich and temporally consistent videos of high fidelity under identical grounding conditions of the depth map. The methods compared include Control-A-Video (CAV)~\cite{chen2023control}, ControlVideo (CV)~\cite{zhang2023controlvideo}, and Gen-1~\cite{esser2023structure}. In the first sample, CV and Gen-1 fail to recognize the face direction of the bear. In the second challenging sample featuring high dynamics motion, other methods tend to generate distorted dog or cannot generate content in the desirable position, while \shortname{} could generate smooth video and the content of each frame is more photo-realistic. The third example reveals that while CAV effectively renders color information, it loses detail in content representation. Gen-1 and CV, influenced by the description ``pink", erroneously colorize the entire scene pink rather than just the targeted object. Besides, \shortname{} could smooth low-quality grounding condition, since we perform temporal attention to the grounding condition. For example, the generated dog in~\cref{fig:main} is even clearer than the dog in the source video. Additional qualitative comparisons are available in the supplementary materials. 
\input{tab/ablation}
\input{tab/comp}
\input{img_tex/skip}

\input{img_tex/long}

\input{img_tex/se}
\input{img_tex/edit}

\noindent\textbf{Quantitative Comparison.} We quantitatively compare \shortname{} and leading methods with the same prompt and depth map grounding conditions as shown in~\cref{tab:comp}. GLIGEN~\cite{li2023gligen} accommodates both continuous and discrete grounding conditions, but falls short in grounded video generation. Other methods exhibit limitations in video generation with discrete grounding conditions while \shortname{} can support both two types of grounding conditions. Further, \shortname{} outperforms the other approaches in both temporal consistency and prompt consistency.

\subsection{Ablation study}
\noindent\textbf{Individual Effectiveness.} To evaluate the effectiveness of the proposed modules, we perform an ablation study on temporal consistency and prompt consistency in~\cref{tab:ab}, consisting of grounding injection, spatial-temporal grounding layer, and dynamic gate network. First, grounding injection could explicitly guide the focus of the visual token integration to improve the performance. With STGL, \shortname{} effectively achieves grounded generation with high temporal and prompt consistency. Further, dynamic gate network adaptively bypasses redundant STGA to selectively extract low-level grounding and high-level semantics, which could slightly improve performance while boosting efficiency. Besides, generating videos with sparse conditions (\eg., layout) could achieve higher quantitative performance, albeit at a coarser granularity. However, we observe that the extracted dense condition like the canny map may be unreliable, which may mislead video generation in the wrong direction.

\noindent\textbf{Dynamic Gate Network.} To validate the efficiency of the proposed dynamic gate network, we analyze the frequency at which each layer is bypassed. The results, illustrated in \cref{fig:skip}, indicate that deeper layers are often skipped. This trend suggests that shallow layers primarily address low-level grounding conditions and deeper layers handle more abstract semantics in downsample blocks. A similar pattern emerges in the upsample blocks. Therefore, DGN could effectively skip the redundant layer to use the corresponding layer to process the corresponding grounding or semantic information while improving efficiency.

\subsection{Applications}

\noindent\textbf{Long-range Generation.} We evaluate \shortname{} on long-range generation via an auto-regressive paradigm as shown in~\cref{fig:long}. To preserve more details, we generate 1200 frames at 30 FPS of complex natural scenery with the grounding conditions of the canny map, which indicates that \shortname{} could effectively generate long-range videos with high fidelity and temporal consistency. Besides, \shortname{} also has the theoretical capability to generate an infinite length of video controlled by grounding conditions.

\noindent\textbf{Non-uniform Sequential Prompts.}  As depicted in~\cref{fig:se}, \shortname{} enables the video generation controlled by sequential prompts, akin to the story-telling mode. The grounded generation capacity with sequential prompt could achieve fine-grained textual-based controllability over video content. Further, prompt interpolation facilitates smooth transitions between specific frames in the generated videos, enhancing the natural flow of the visual narrative.

\noindent\textbf{Object-specific Editing.} The effectiveness of object-specific editing using \shortname{} is showcased in~\cref{fig:edit}.  \shortname{} effectively edits the specific object via text prompt into SAM, \ie., car and human into the anime style car and Iron Man while preserving the overall background in high consistency. These two challenging scenarios are accompanied by high dynamic motion, which further validates the effectiveness of \shortname{}.

%% file: img_tex/main.tex
 \begin{figure*}[ht]
    \begin{center}
       \includegraphics[width=\textwidth]{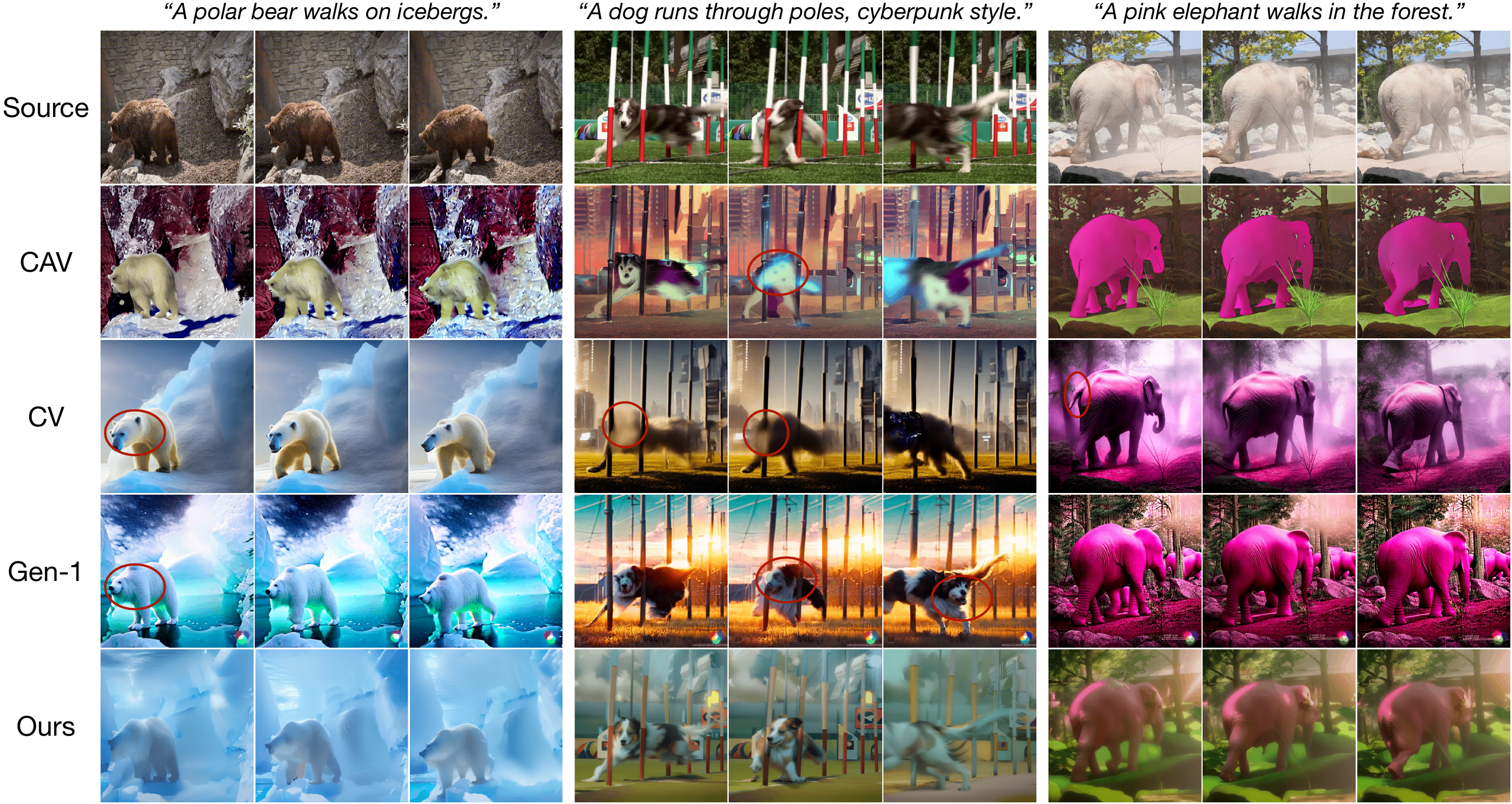}
    \end{center}
    \caption{Comparison with state-of-the-art-methods, including Control-A-Videl (CAV)~\cite{chen2023control}, ControlVideo (CV)~\cite{zhang2023controlvideo}, and Gen-1~\cite{esser2023structure}.}
    \label{fig:main}
 \end{figure*}

%% file: tab/ground.tex
\begin{table}[t]
\centering
\setlength{\tabcolsep}{0.6mm}
\caption{Quantitative comparison of grounded generation capacity.}
\begin{tabular}{c|ccccc}
\toprule
Method & Layout(AP) & Depth & Normal & HED & Canny \\ \midrule
Control-A-Video     & --         &  96.7     &    96.0    & 96.3    &   96.5    \\
ControlVideo    & --         &   97.3    &  96.7      & 97.1    &  96.9     \\
Ours   &   \textbf{26.3}         &    \textbf{99.0}   &    \textbf{98.1}    & \textbf{98.6}   &    \textbf{98.2}   \\ \bottomrule
\end{tabular}
\label{tab:grounded}
\end{table}

%% file: tab/ablation.tex
\begin{table*}[t]
\centering
\small
\caption{Ablation of \shortname{} framework, including uncertainty-based grounding injection (GJ), spatial-temporal grounding layer (STGL), and dynamic gate network (DGN).}
\begin{tabular}{ccc|cccc|cccc}\toprule
\multirow{2}{*}{GJ} & \multirow{2}{*}{STGL} & \multirow{2}{*}{DGN} & \multicolumn{4}{c|}{Temporal Consistency} & \multicolumn{4}{c}{Prompt Consistency} \\ \cmidrule{4-11}
                    &                      &                                  & Layout    & Canny    & Depth    & Norm   & Layout    & Canny   & Depth   & Norm   \\ \midrule
           &                      &                                  & 79.83 &  78.76 & 78.95 & 77.40 & 24.68 & 22.29 & 24.48 & 21.71 \\ 
\usym{1F5F8}            &                      &                                  &  83.47         & 82.13         &     82.51     &    81.06    &  24.91         &   22.54      &  24.88       &   22.18     \\
\usym{1F5F8}          &         \usym{1F5F8}             &                                  &  98.02         &   96.94       &    97.45      &  96.30      &    26.16       &       24.91  &  26.82       &    24.49   \\
\usym{1F5F8}         &       \usym{1F5F8}               &         \usym{1F5F8}                         &      \textbf{98.33}     &   \textbf{97.42}       &   \textbf{97.83}       &  \textbf{96.84}      &     \textbf{26.84}      &    \textbf{25.39}     &  \textbf{27.07}       &  \textbf{24.78}       \\
\bottomrule
\end{tabular}
\label{tab:ab}
\end{table*}

%% file: tab/comp.tex
\begin{table}[t]
\centering
\small
\caption{Quantitative comparison between preceding methods and  \shortname{} on temporal consistency and prompt consistency.}

\setlength{\tabcolsep}{0.6mm}
\begin{tabular}{c|cc|c|cc}
\toprule
\multirow{2}{*}{Method} & \multicolumn{2}{c|}{Grounding Condition}                            & \multirow{2}{*}{Video} & \multirow{2}{*}{TC.} & \multirow{2}{*}{PC.} \\ \cmidrule{2-3}
                        & \multicolumn{1}{c|}{Discrete} & \multicolumn{1}{c|}{Continuous} &                        &                            &                               \\ \midrule
GLIGEN                  & \multicolumn{1}{c|}{\usym{1F5F8}}         &       \usym{1F5F8}                         &        --                &                      78.95    &   24.48                      \\
Control-A-Video         & \multicolumn{1}{c|}{--}       &            \usym{1F5F8}                 &       \usym{1F5F8}                 &                      94.67      &  23.08                      \\
ControlVideo & \multicolumn{1}{c|}{--}   & \usym{1F5F8}& \usym{1F5F8}& 96.91 & 25.28 \\ 
Gen-1 & \multicolumn{1}{c|}{--}   & \usym{1F5F8}& \usym{1F5F8}& 94.57 & 25.95 \\
Ours                    & \multicolumn{1}{c|}{\usym{1F5F8}   }         & \usym{1F5F8}                                  &  \usym{1F5F8}                         &      \textbf{97.83}                      &    \textbf{27.07}               \\
\bottomrule
\end{tabular}
\label{tab:comp}
\end{table}

%% file: img_tex/skip.tex
 \begin{figure}[t]
    \begin{center}
       \includegraphics[width=0.48\textwidth]{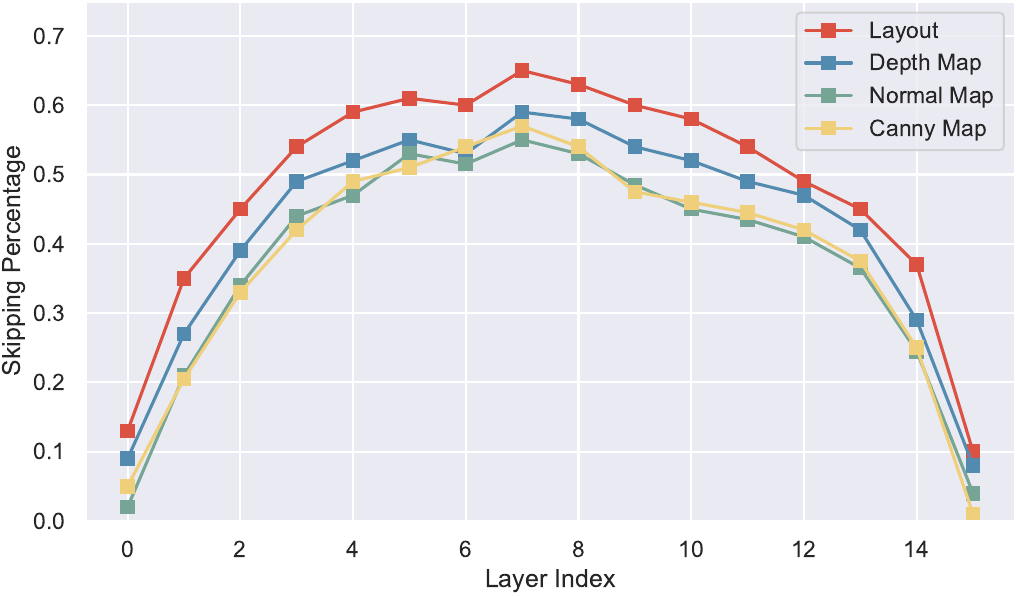}
    \end{center}
    \caption{Skipping percentage of each layer with Dynamic Gate Network (DGN).}
    \label{fig:skip}
 \end{figure}

%% file: img_tex/long.tex
 \begin{figure*}[h]
    \begin{center}
       \includegraphics[width=0.98\textwidth]{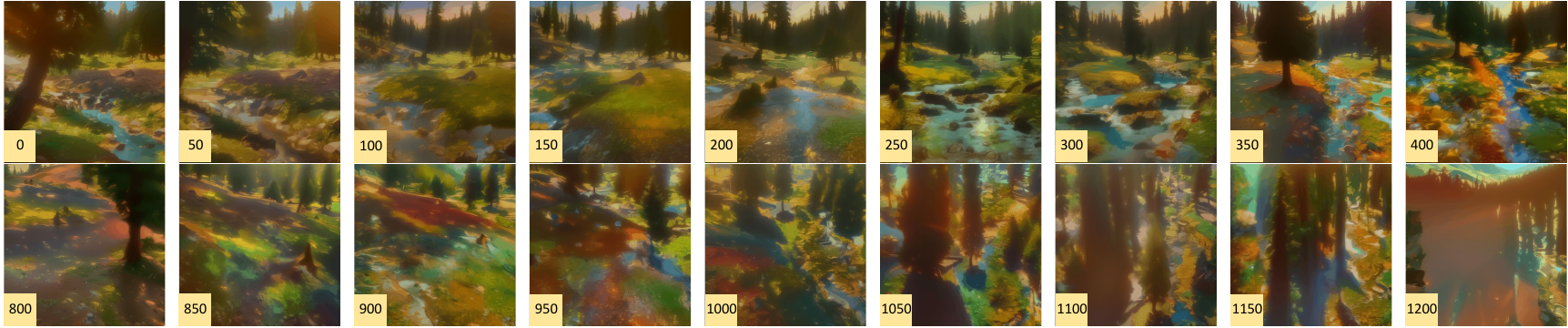}
    \end{center}
    \caption{Results of long-range video generation conditioned on canny map. \textit{Prompt}: A natural landscape of mountains and jungle on a sunny day, as viewed from a drone.}
    \label{fig:long}
 \end{figure*}

%% file: img_tex/se.tex
\begin{figure*}[t]
\centering
	\subcaptionbox{Change blue autumn clothing into red down jacket started at 4$^{th}$ frame.} {\includegraphics[width = 0.98\textwidth]{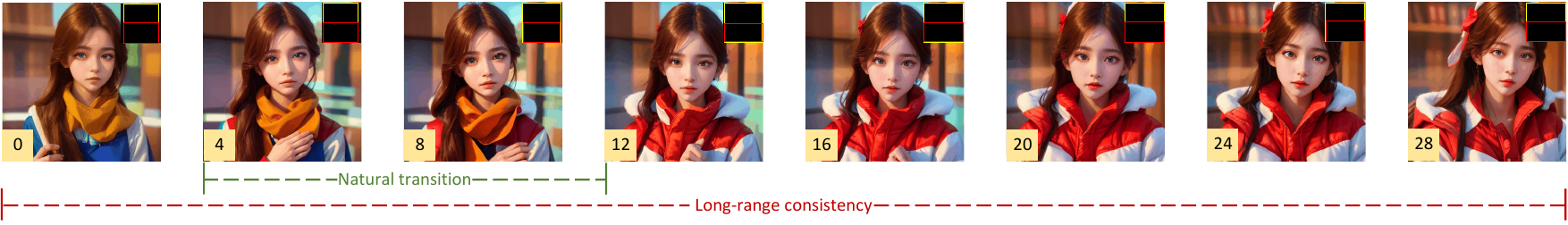}}
	\hspace{0.3cm}
	\subcaptionbox{Change furry suit and black pants of a humanoid cat into a sharp suit with white pants started at 16$^{th}$ frame. }{\includegraphics[width = 0.98\textwidth]{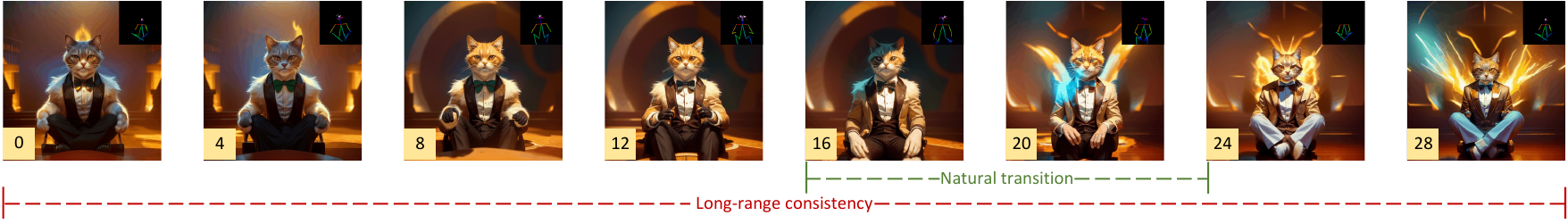}}
        
\caption{Results of the grounded T2V generation with non-uniform sequential prompts, which are simplified into natural language for better understanding.}
\label{fig:se}
\end{figure*}

%% file: img_tex/edit.tex
 \begin{figure}[ht]
    \begin{center}
       \includegraphics[width=0.49\textwidth]{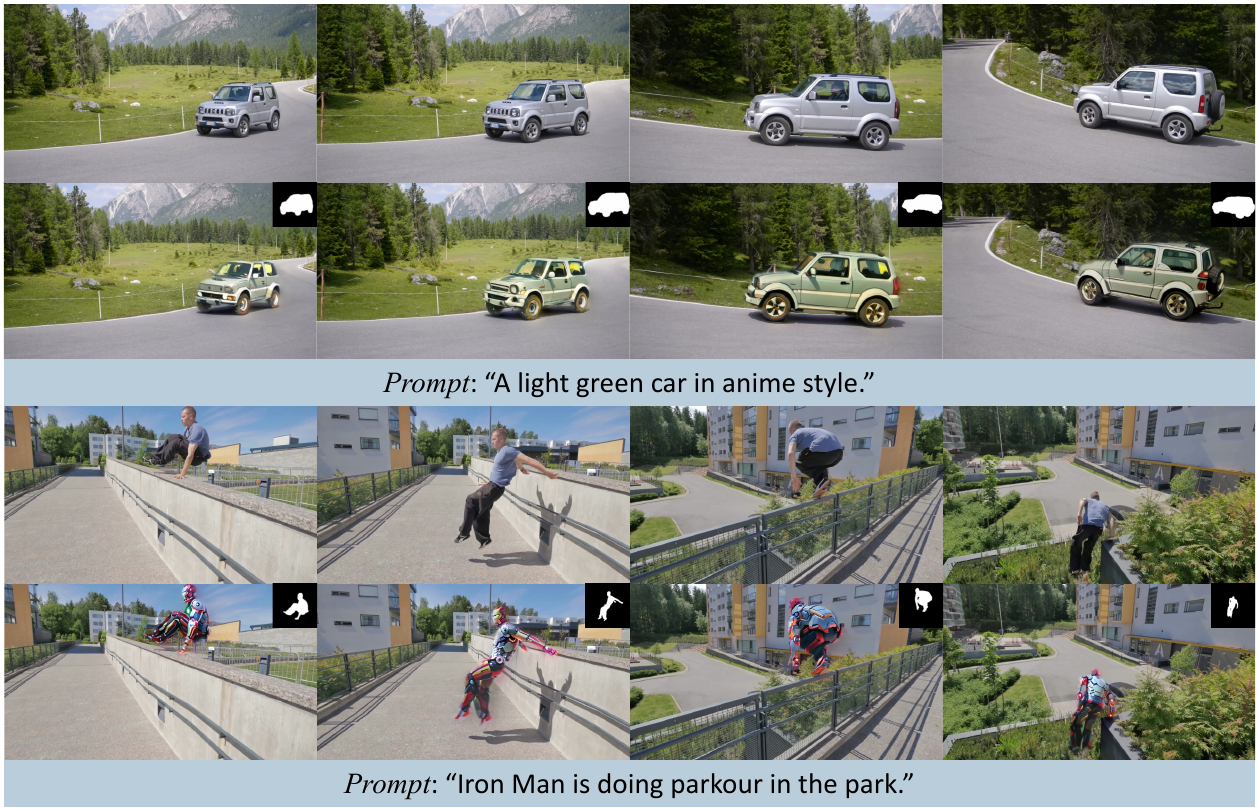}
    \end{center}
    \caption{Results of object-specific editing, which edits the target object while maintaining the background fixed.}
    \label{fig:edit}
 \end{figure}

%% file: sec/5_conclusion.tex
\section{Conclusion and Limitations}

In this paper, we present the Grounded Text-to-Video Generation framework (\shortname{}), enabling grounded video generation capacity under both discrete and continuous grounding conditions. The proposed grounding injection explicitly provides the grounding prior to guide the focus region of the network. Spatial-temporal grounding layer facilitates grounded generation in the spatial-temporal domain. Further, our dynamic gate network adaptively skips the unnecessary grounding module to selectively extract low-level grounding and high-level semantics while improving efficiency. Extensive experiments validate the ability of \shortname{} with promising grounded generation, temporal, and prompt consistency. Based on the grounded generation ability, we extend \shortname{} on three practical applications, including long-range video generation, non-uniform sequential prompts, and object-specific editing.

We also observe some failure cases. A major challenge for \shortname{} is understanding complex interactions between objects. This issue partly stems from the coarse-grained text-image alignment of the CLIP training~\cite{radford2021learning}. We hypothesize that integrating a stronger language model like LLM or recaption technology~\cite{zeqiang2023mini} may mitigate this issue.

%% file: sec/comp1.tex
 \begin{figure*}[h]
    \begin{center}
       \includegraphics[width=\textwidth]{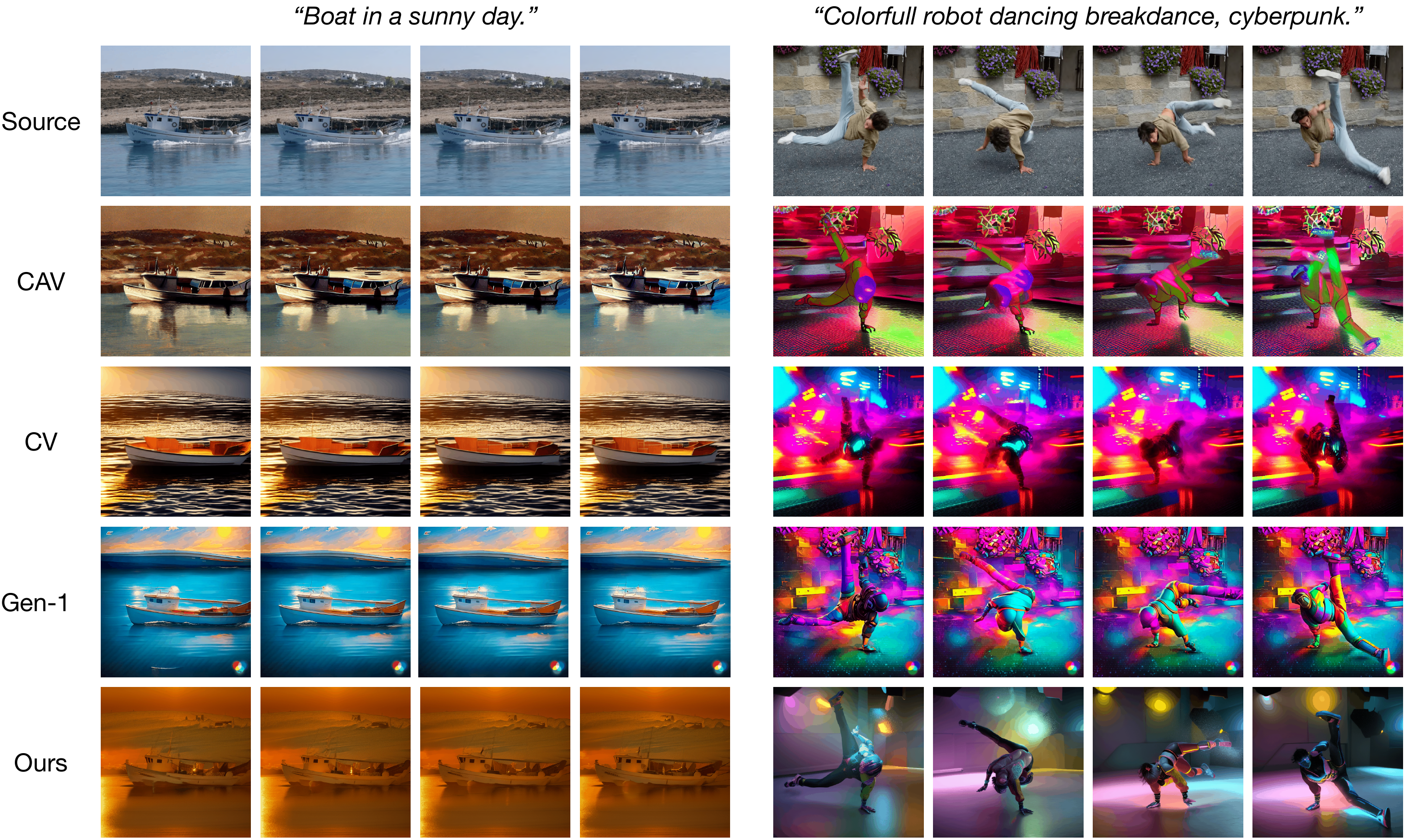}
    \end{center}
    \caption{Comparison with previous methods.}
    \label{fig:1}
 \end{figure*}
 \begin{figure*}[h]
    \begin{center}
       \includegraphics[width=\textwidth]{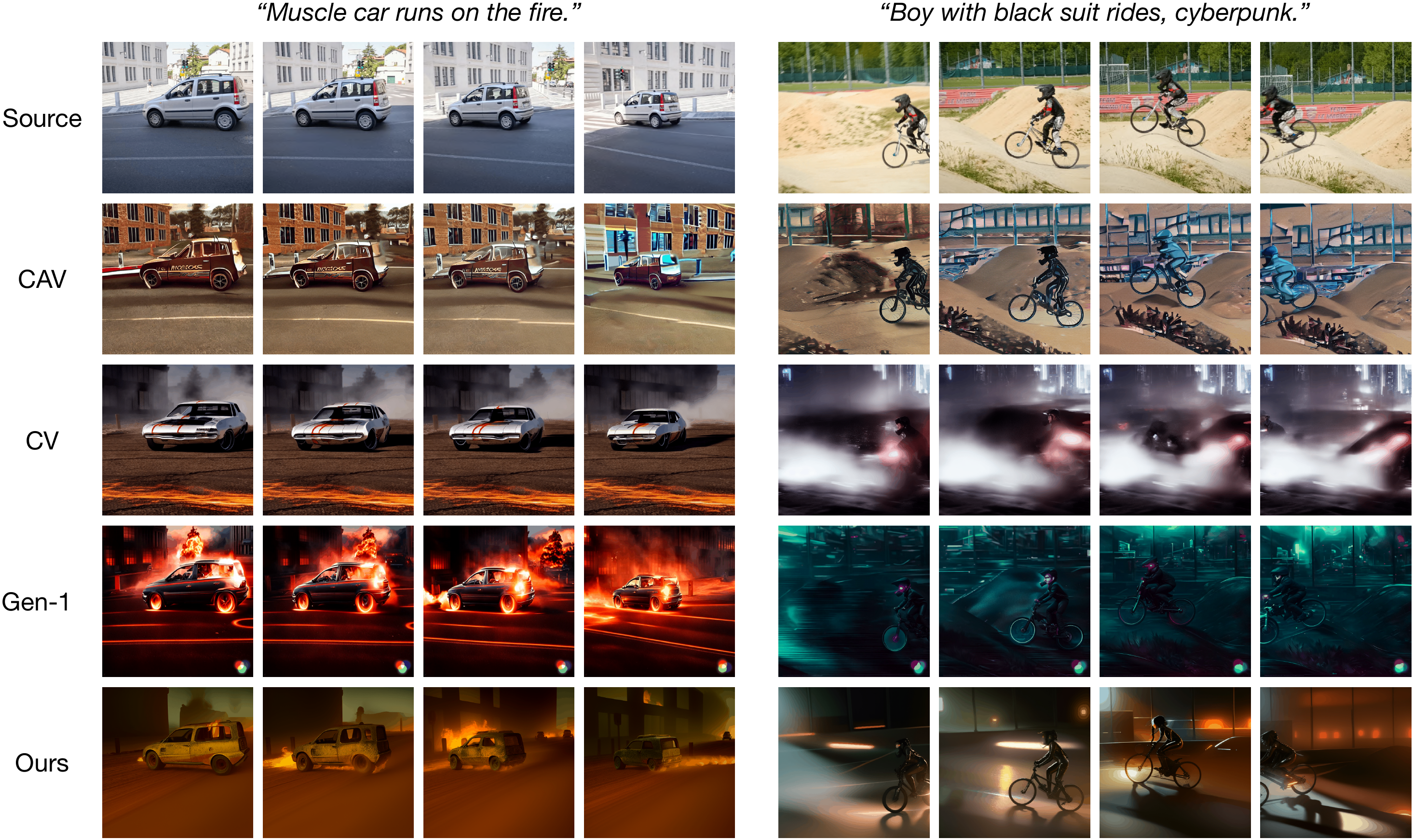}
    \end{center}
    \caption{Comparison with previous methods.}
    \label{fig:2}
 \end{figure*}
 \begin{figure*}[h]
    \begin{center}
       \includegraphics[width=\textwidth]{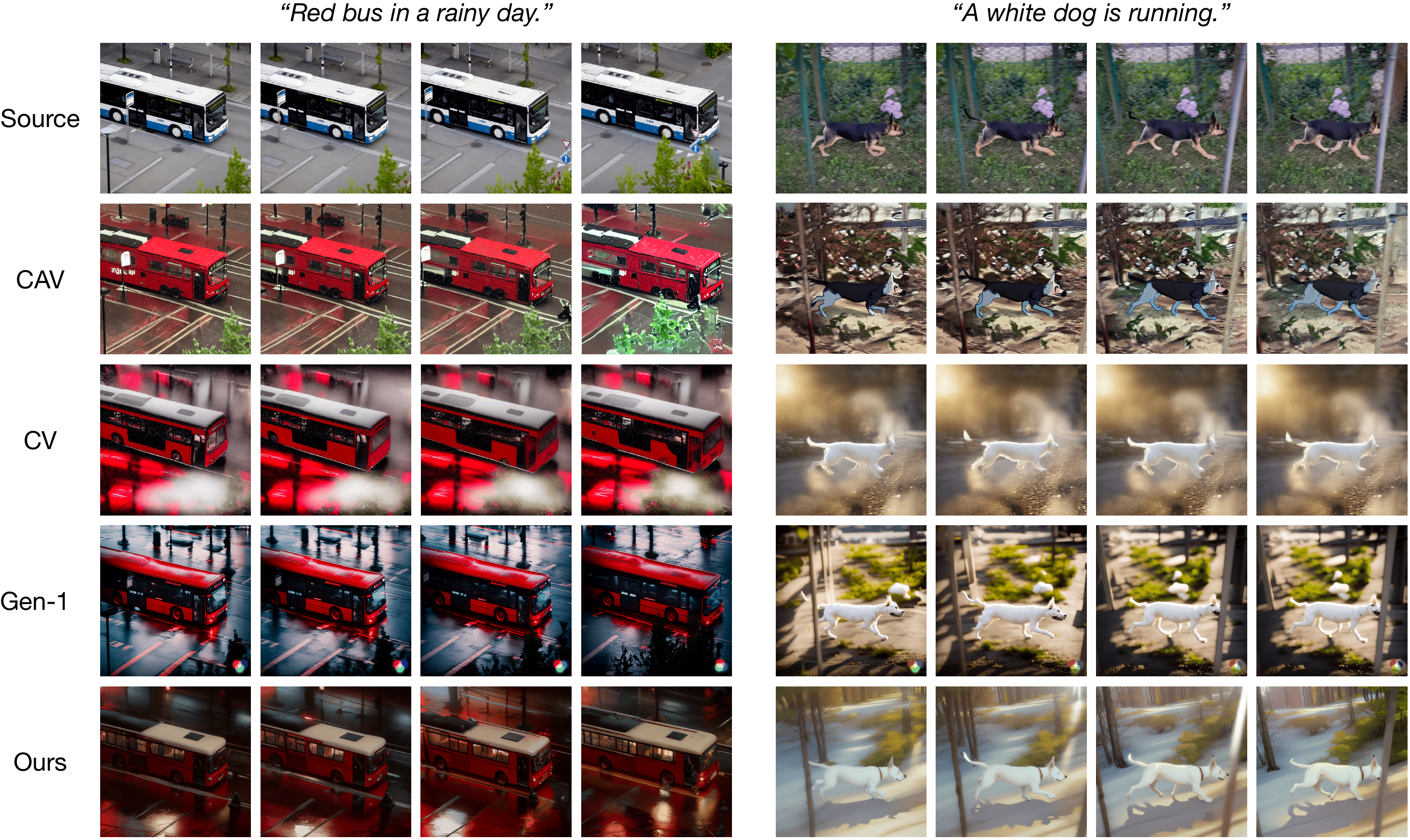}
    \end{center}
    \caption{Comparison with previous methods.}
    \label{fig:3}
 \end{figure*}
 \begin{figure*}[h]
    \begin{center}
       \includegraphics[width=\textwidth]{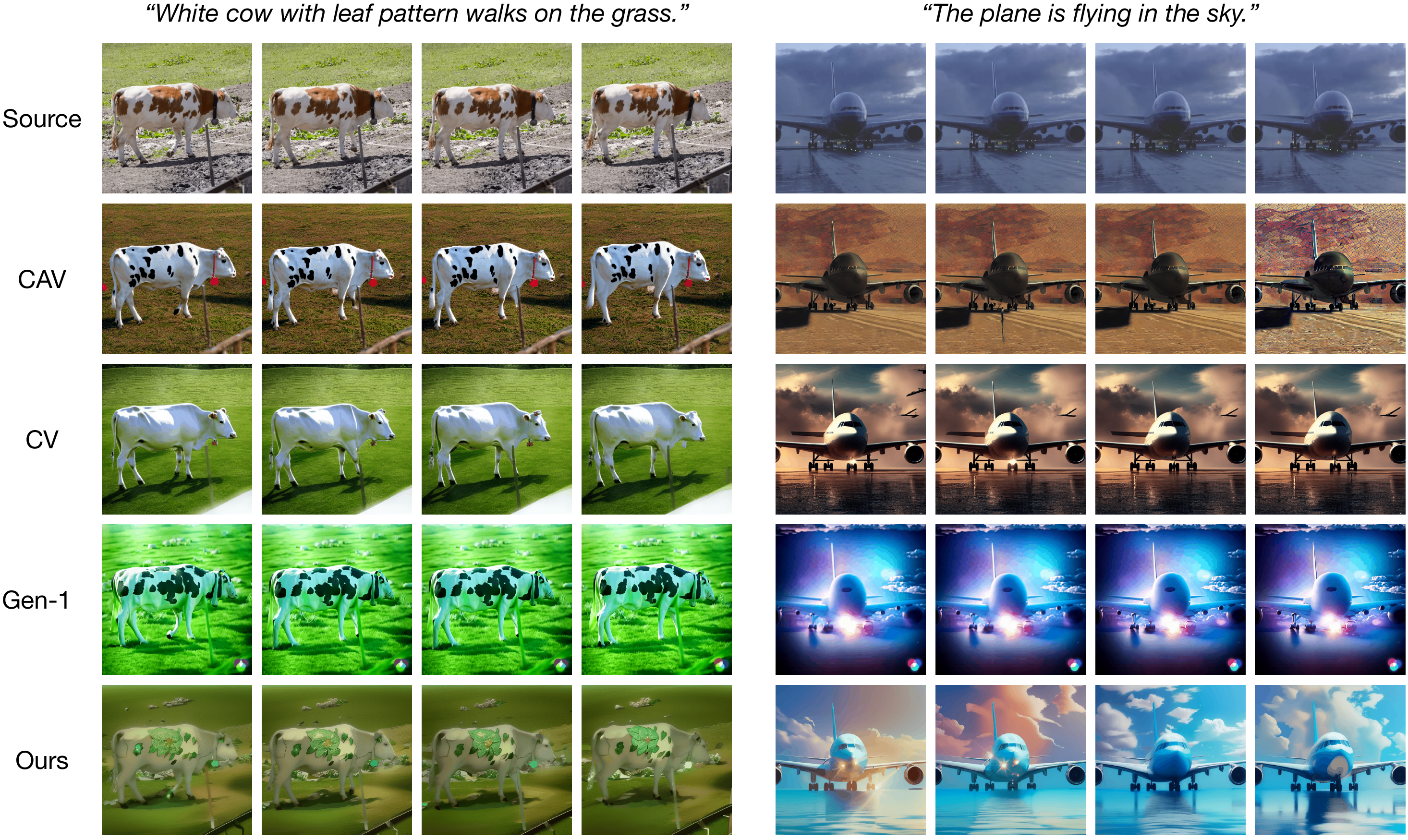}
    \end{center}
    \caption{Comparison with previous methods.}
    \label{fig:4}
 \end{figure*}
 \begin{figure*}[h]
    \begin{center}
       \includegraphics[width=\textwidth]{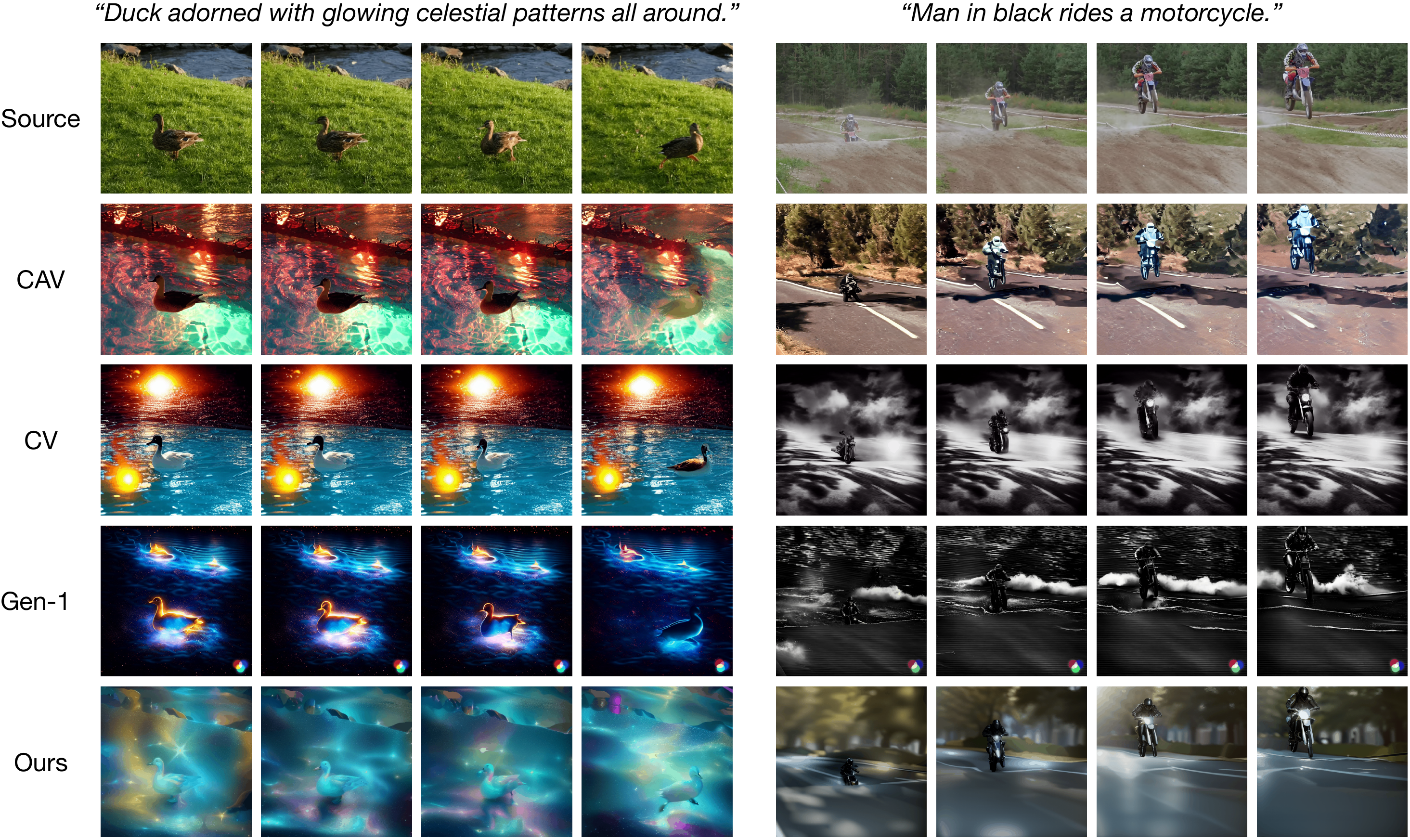}
    \end{center}
    \caption{Comparison with previous methods.}
    \label{fig:5}
 \end{figure*}

%% file: sec/per.tex
\begin{figure*}[t]
\centering
	\subcaptionbox{Results with mistoonAnime.} {\includegraphics[width =0.8\textwidth]{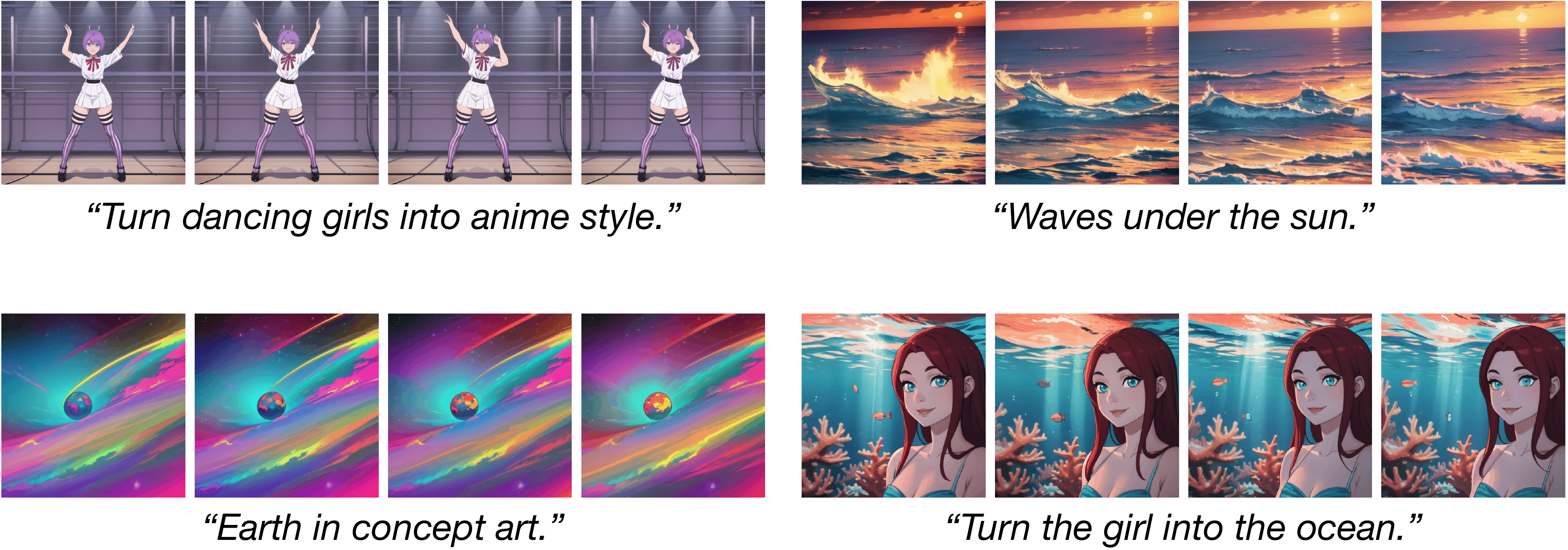}}
	
	\subcaptionbox{Results with realisticVision. }{\includegraphics[width =0.8\textwidth]{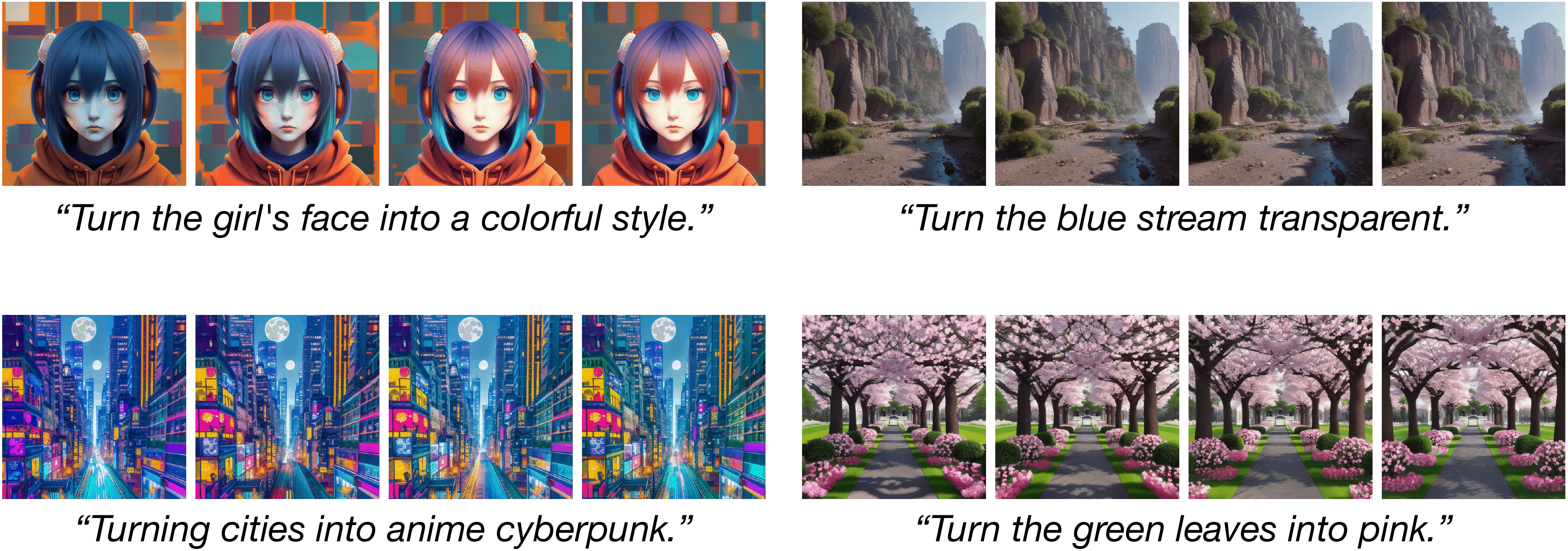}}
 \subcaptionbox{Results with majicmixRealistic. }{\includegraphics[width = 0.8\textwidth]{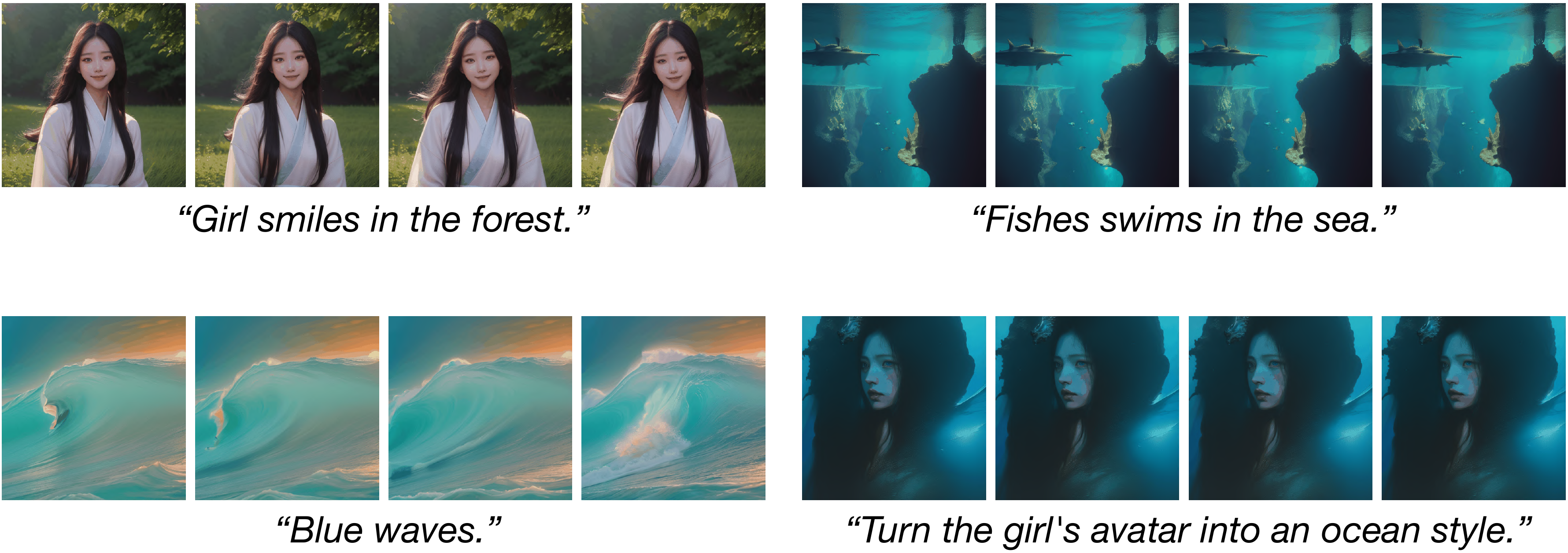}}
 \subcaptionbox{Result with xxmix9realistic. }{\includegraphics[width = 0.8\textwidth]{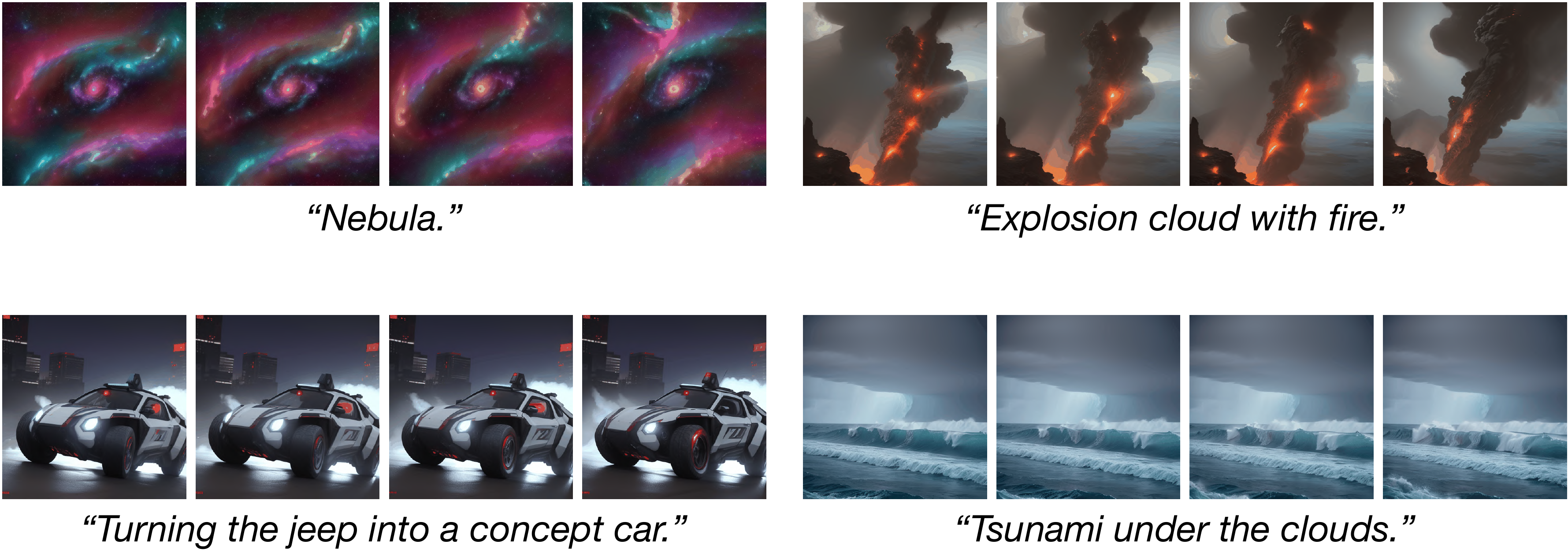}}
        
\caption{Results conditioned on web videos with popular personalized models from CivitAI~\cite{civitai}.}
\label{fig:per}
\end{figure*}
\clearpage